\documentclass[preprint,12pt]{elsarticle}

\usepackage{amssymb}
\usepackage{amsmath}

\usepackage{makecell}
\usepackage[ruled]{algorithm2e}
\usepackage{graphicx}
\usepackage{subfigure}
\usepackage{float}
\usepackage{threeparttable}
\usepackage{tabularx}
\usepackage{amsmath}
\usepackage{multirow}

\usepackage{soul}
\usepackage{bm}

\usepackage{color, xcolor}

\journal{Neurocomputing}

\begin{document}

\begin{frontmatter}



\title{A Stock Price Prediction Approach Based on Time Series Decomposition and Multi-Scale CNN using OHLCT Images}


\author[must]{Zhiyuan Pei}
\affiliation[must]{
            organization={School of Computer Science and Engineering, Faculty of Innovation Engineering, Macau University of Science and Technology},
            city={Macau},
            postcode={999078},
            country={China}}

\affiliation[lx]{
            organization={Macau Institute of Systems Engineering, Faculty of Innovation Engineering,  Macau University of Science and Technology},
            city={Macau},
            postcode={999078},
            country={China}}

\author[must]{Jianqi Yan}

\author[must]{Jin Yan}

\author[must]{Bailing Yang}

\author[li]{Ziyuan Li}
\affiliation[li]{
            organization={Emerging Technology, Innovation, and Ventures, HSBC},
            city={Shenzhen},
            postcode={518052},
            country={China}}

\author[zhang]{Lin Zhang}
\affiliation[zhang]{
            organization={School of Accounting and Finance, Beijing Institute of Technology},
            city={Zhuhai},
            postcode={519088},
            country={China}}

\author[lx]{Xin Liu\corref{cor1}}
\cortext[cor1]{Corresponding author: xiliu@must.edu.mo (Xin Liu)}

\author[zhang2]{Yang Zhang}
\affiliation[zhang2]{
            organization={Department of Mathematics, University of Manitoba},
            city={Winnipeg},
            postcode={R3T 2N2},
            country={Canada}}

\begin{abstract}

Recently, deep learning in stock prediction has become an important branch.
Image-based methods show potential by capturing complex visual patterns and spatial correlations, offering advantages in interpretability over time series models. 
However, image-based approaches are more prone to overfitting, hindering robust predictive performance.
To improve accuracy, this paper proposes a novel method, named Sequence-based Multi-scale Fusion Regression Convolutional Neural Network (SMSFR-CNN), for predicting stock price movements in the  China  A-share market.

Firstly, the historical opening, highest, lowest, closing price, and turnover rate (OHLCT) of stocks are converted into images, separated by weekends, with time information to help the CNN learn the impact of different trading periods.
To reduce overfitting, long sequences of stock features are decomposed into multiple time periods,  and OHLCT images at different time scales are utilized as inputs, significantly reducing overfitting.
Thirdly, in order to overcome the problem that classification labels lose information about the magnitude of stock price changes, we introduce regression labels to help the model capture more latent features of stock price fluctuations.

By utilizing CNN to learn sequential features and combining them with image features, we improve the accuracy of stock trend prediction on the A-share market stock dataset.
This approach reduces the search space for image features, stabilizes and accelerates the training process.
Extensive comparative experiments on 4,454 A-share stocks show that the model achieves 61.15\% positive predictive value and 63.37\% negative predictive value for the next 5 days, resulting in a total profit of 165.09\%.

\end{abstract}

\end{frontmatter}

\section{Introduction}

With the rapid development of financial markets and the acceleration of globalization, stock investment has become an essential avenue for many investors to achieve wealth appreciation \cite{yuniningsih2017analysis, mintarya2023machine, vullam2023prediction}. However, the complexity and uncertainty of the stock market make accurately predicting stock trends a challenging task. The China A-share market, being one of the largest and most dynamic globally, has attracted significant quantitative analysis and research due to the growing impact of economic globalization \cite{wang2007cross, faxing2024green, qian2024implementation}.

Traditional stock analysis methods heavily reliant on financial data and macroeconomic indicators and often fall short in capturing the dynamic and non-linear relationships within the market \cite{alam2013forecasting, banerjee2014forecasting, dana2016modelling, mallikarjuna2019comparison}. Therefore, quantitative trading has emerged as a pivotal component of modern financial strategies. By leveraging computational techniques, it enables the execution of trading strategies devoid of emotion-based decision-making, thereby uncovering patterns that may elude human analysts \citep{ferreira2021analysis}. The potential  quantitative trading in stock market forecasting has been well-documented \citep{carpenter2021real}. This progress has led to the development of advanced deep models that can handle the inherent complexity, nonlinearity, and noise of financial markets, making the construction of stock trend prediction models using machine learning, deep learning, and big data techniques, which becomes a hot research topic in finance \cite{shen2012stock, leung2014machine,  ma2021parallel, patel2015predicting, leippold2022machine, sonkavde2023forecasting, cui2023mcvcsb}.

Despite significant advancements, many deep learning models for stock price prediction still grapple with challenges such as data noise, model overfitting, and insufficient interpretability \citep{he2016deep}. 
Baek et al. \cite{baek2018modaugnet} pointed out that the limited number of training data points often leads to overfitting in deep neural network models. Ito et al. \cite{ito2021trader} aggregated several simple, interpretable weak algorithms and assessed the importance of each weak algorithm in improving overall interpretability.
In order to tackle the data noise problem, the research conducted by Liu et al. \cite{liu2019stock} employed sparse autoencoders with one-dimensional (1D) residual convolutional networks to denoise the stock prices.

Convolutional neural networks (CNNs) have been employed for stock market prediction due to their efficient feature extraction capabilities \citep{hoseinzade2019cnnpred, lu2021cnn, mehtab2020stock}.
However, traditional CNN approaches often fail to fully utilize time series information, resulting in suboptimal performance under complex market conditions \citep{gunduz2017intraday}. Additionally, when considering multiple stocks, these models typically focus on individual stock information and neglect  the inter-stock correlations that significantly influence price fluctuations \citep{chen2021novel}.
Jiang et al. \cite{jiang2023re} utilized the opening prices, highest price, lowest price, closing price, and volume (OHLCV) images as inputs to predict the probability of stock price increasing  or decreasing, achieving a significant accuracy.
Their experimental results showed that 5-day feature maps in CNNs yield significantly better accuracy than 20-day and 60-day maps, suggesting that longer time series of stock feature maps make it challenging for CNN models to identify critical feature points, leading to local feature overfitting.
Converting sequences to image features can lose the advantage of utilizing more historical data, whereas shorter image features risk sacrificing significant historical information, increasing prediction inaccuracies.

Inspired by these studies, we first propose two novel methods.
The first method involves replacing trading volume with stock turnover rate, which eliminates the impact of trading volume caused by ex-rights events, resulting in more stable features.
The second method introduces an integration between time separator and OHLCT (Opening price, Highest price, Lowest price, Closing price, and Turnover rate), resulting a new image feature as input, named as TS-OHLCT.
Specifically, we incorporate the weekend time information as separators into OHLCT images to help CNNs capture trading temporal information and learn the effects of stock trading cycles.

Furthermore, two new architectures are proposed.
The first is a Multi-Scale Residual Convolutional Neural Network (MSR-CNN) designed to address the overfitting problem in long sequence images.

The second architecture is a Sequence-based Multi-scale Fusion Regression Convolutional Neural Network (SMSFR-CNN), which addresses the problem that using only image features makes it difficult to learn information about stock price fluctuations. By integrating sequence data, SMSFR-CNN better captures stock price trends in the A-share market.
We observed that traditional CNN methods often exhibit local feature overfitting and convergence issues in stock image feature extraction, with prediction performance gradually deteriorating as the time of sequence feature maps increases.
This observation aligns with the findings of \cite{jiang2023re}.
To address these mentioned issues, we decomposed long sequences of stock image features into multiple time periods according to different time scales and assign different feature weights based on their importance (with higher weights for features closer to the current trading day).
This significantly reduces overfitting and enables CNNs to  learn long sequence image features better.

Additionally, considering that investors  more concern about  the magnitude of price fluctuations rather than  the probability of price increasing  or decreasing, and given that it is difficult for the image features to effectively capture the magnitude of price changes,   the proposed model integrates time series information as an extra features.
Our approach utilizes the CNN component to learn time series features and concatenates them with image features, simultaneously predicting both the magnitude and probability of stock price movements.
This method effectively incorporates regression labels into the existing framework.
Our experimental results indicate that this approach significantly improves the accuracy of stock price trend predictions, reduces the search space for image features,  stabilizes and accelerates the convergence process.

 The paper's major contributions can be summarized as follows:

\begin{itemize}
\item This is the first time that historical open, high, low, close prices and turnover-rates are incorporated into images, which are separated by weekends and combined with time-specific information.
\item The long sequences of stock image features are decomposed into multiple time periods according to different time scales and assigned different feature weights based on their importance (with higher weights for features closer to the current trading day).
\item Combining time series  with image features using CNN improves prediction accuracy, reduces the search space,  stabilizes and accelerates the convergence process.
\item Comprehensive comparison experiments between different methods on 4,454 A-share stocks are provided, and the majority of A-share stocks are considered in our experiments.
\item Our proposed method, SMSFR-CNN, outperforms other advanced methods in terms of positive predictive value (PPV) and negative predictive value (NPV).

\end{itemize}

The remainder of the paper is structured as follows. The related works of CNN methods in stock prediction are introduced in Section \ref{sec2}.
Section \ref{sec3} presents the dataset that serves as the basis for our study along with a detailed description of the innovative and predictive model developed in this paper.  In Section \ref{sec4}, we describe  and analyze the experimental settings and results.
Finally,  we provide  a summary of the conclusions and main contributions of the paper in Section \ref{sec5}. We also highlight the significance of our results and propose potential works for future research.

\section{Related Work}\label{sec2}
The application of CNNs in predicting stock prices has gained  significant attention  for their capacity to capture complex patterns in time-series data.
Early approaches primarily relied on traditional machine learning techniques.
However, the emergence of deep learning  brought innovative  ways to handle the inherent non-linearities and temporal dependencies in financial data.

The use of CNNs in stock price prediction was motivated by their success in image recognition tasks, where spatial hierarchies of features are learned.
This concept was extended to financial data by treating time-series data as a form of sequential image data, allowing CNNs to extract meaningful features from raw inputs. 
Aadhitya et al. \cite{bagde2023predicting} highlighted how CNNs could be employed to capture the short-term patterns in stock data, which were crucial for making accurate predictions.
By combining CNNs with the Long Short-Term Memory (LSTM) networks, their model leveraged both spatial and temporal features, and thus improved  the prediction accuracy compared to traditional models.
Another significant contribution was given by  Hoseinzade et al. \cite{hoseinzade2019cnnpred}, where the authors proposed a CNN-based model that incorporated a wide range of financial indicators.
The study demonstrated that CNNs could effectively process and learn from diverse data representations, enhancing the robustness of stock price predictions.
The model's ability to integrate different data sources was particularly advantageous in the volatile stock market environment, where multiple factors simultaneously influenced stock prices.

Several studies proposed enhancements to the basic CNN architecture to better suit the unique characteristics of financial data.
For instance, Hoseinzade et al. \cite{hoseinzade2019u} proposed a universal CNN framework that adapted to various stock markets by tuning the network's hyperparameters.
This adaptability was crucial for deploying models across different markets with varying levels of volatility and liquidity.

Moreover, the integration of CNNs with other deep learning models was explored to address the limitations of standalone CNNs.
The work by Kim et al. \cite{kim2019forecasting} combined the strengths of CNNs and LSTMs allow  the model to capture both local dependencies in stock prices and long-term trends. This hybrid model demonstrated superior performance in predicting stock prices over different time horizons, showcasing the synergy between CNNs and recurrent architectures.

A novel application of CNNs was in the analysis of candlestick charts.
The study by Hu et al. \cite{hu2018deep} employed CNNs to automatically learn features from candlestick chart images, enabling the model to make informed investment decisions.
This approach automated the traditionally manual task of chart analysis and also enhanced the accuracy of predictions by capturing intricate patterns that might have been overlooked by human analysts.
Similarly, the work by Rosdyana et al. \cite{kusuma2019using} further validated the effectiveness of CNNs in analyzing visual representations of stock data.
The study illustrated how CNNs could be trained to recognize specific candlestick patterns, which were then used to predict future stock movements.
Their method highlighted the potential of CNNs to transform visual financial analysis into a data-driven, automated process.

While CNNs showed great promise in stock price prediction, several challenges remained.
The primary concern was the model's capability to generalize across various market conditions, which could vary drastically over time.
Furthermore, the high dimensionality of financial data posed a risk of overfitting, particularly when the training data was limited.
In our work, we proposed a novel image format with various features into an image, allowing CNNs learn the open, close, highest, lowest prices, turnover rates and etc. from the images.
To further handle the local overfitting problem, the proposed multi-scale resolution CNN method learns features from long-term stock sequences and decomposes the entire feature map to several sub-feature maps.

\section{Materials and Methods}\label{sec3}

\subsection{Data Modalities}\label{sec3.1}

In this part, we primarily focus on presenting the various data modalities utilized in our framework and introducing a novel data format employed in our experiments.
Using time series as feature inputs in deep models is crucial and widely used in stock market analysis and forecasting.
By integrating several features, including fundamental statistical characteristics, daily returns, price volatility, SMA results, trading volume, transaction volume and lag feature, a set of multivariate time series are obtained as the time series input features for the proposed model.

In addition, many image formats as inputs for deep models are commonly utilized in quantitative trading, including candlestick charts \citep{kusuma2019using, chen2020encoding}, line charts \citep{luo2023learned}, the Open-High-Low-Close-Volume (OHLCV) charts \citep{jiang2023re, lu2022can}, etc.
Some researches constructed the images for convolutional neural networks (CNNs) by applying the sliding window approach to time series \citep{gudelek2017deep}.
Various image formats are employed to depict crucial information in the stock data and act as input features for the deep learning model.
This demonstrates the efficacy and immenses possibilities of utilizing visuals.

\begin{figure*}[htp]
	\centering
	\includegraphics[scale=0.7]{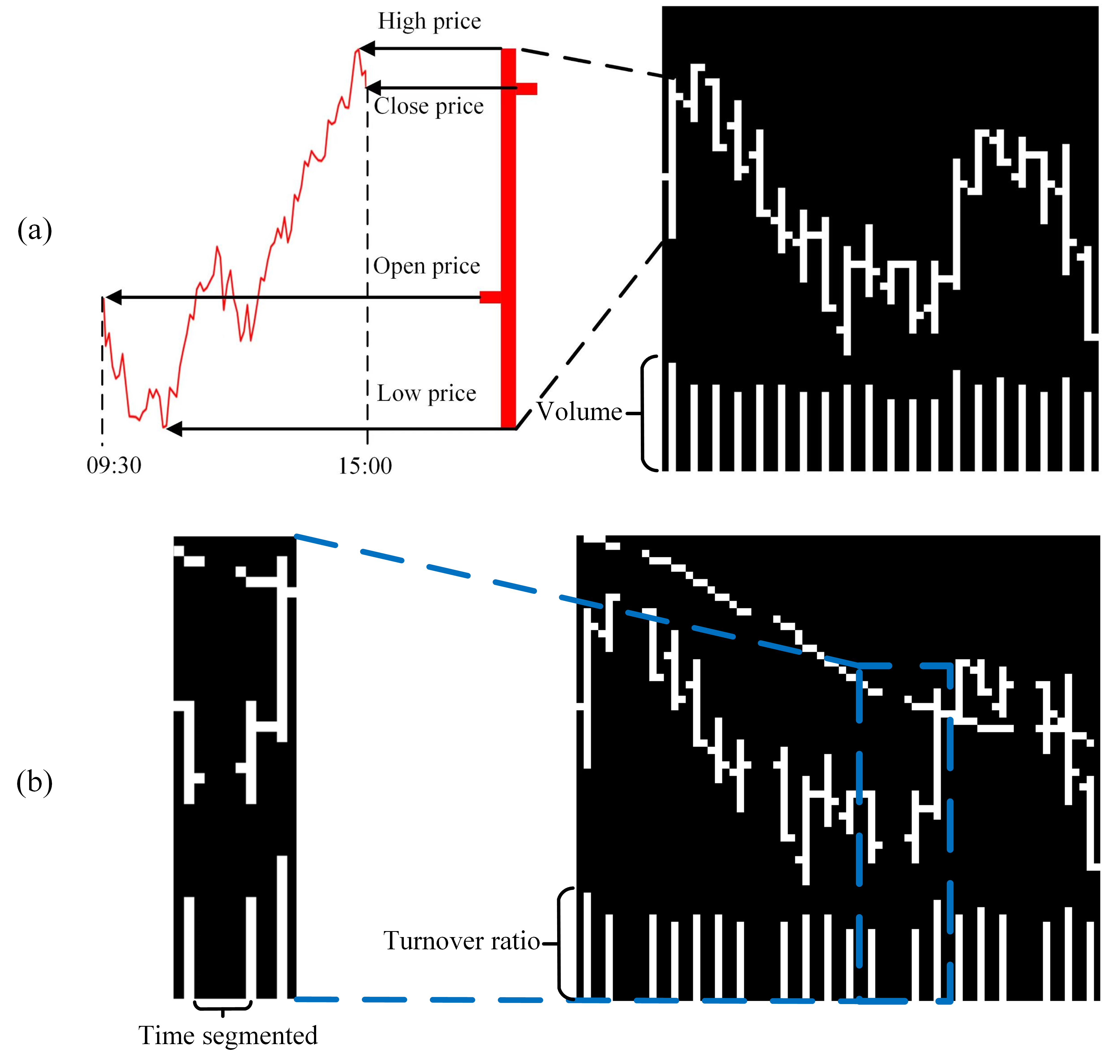}
	\caption{Illustrations of the OHLCV and TS-OHLCT chart in (a) and (b), respectively.}
	\label{fig1}
\end{figure*}

The illustration of the OHLCV chart is presented in Figure \ref{fig1} (a). The opening and closing prices  represent the prices of the first and  last trades 
 within a specific time period, typically a trading day or time frame.
Both opening and closing prices are typically shown as a horizontal line segment on the left side and right side of the vertical line in the OHLC chart.
The highest and lowest trading prices represent the maximum and minimum prices attained by the stock or asset during the same time frame, marking as the upper and lower ends of a vertical line segment in the OHLC chart.
The OHLC chart is organized in chronological order to depict a certain time interval.

In order to enhance the information displayed in an OHLC chart, it is common practice to incorporate moving average (MA) lines and trading volume in the OHLCV charts, as mentioned in \citep{jiang2023re}.
This chart format is beneficial for observing price fluctuations, market trends, trading volume and the intensity of trading activities. 
Inspired by their work, we propose a novel image data format called Time Segmented Open-High-Low-Close Turnover-rate Chart (TS-OHLCT), which is an enhanced version of OHLCV.

We give the main reason for using the turnover rate feature to replace the volume feature.
Adjustment has an impact on the volume feature of stock trading in the A-share market.
Adjustment refers to the procedure of adjusting historical price data to reflect the impact of corporate actions (e.g., dividends, stock bonuses, stock placements, etc.) on the share prices.
Volumes also need to be adjusted for the adjustment ratios, and the adjusted volume data make direct comparisons between different time points or different stocks difficult.
Therefore, given the historical trading volume problem, the concept of utilizing alternative attributes in place of trading volume has emerged.
Our findings indicate that substituting turnover rate for trading volume yields superior outcomes.
Turnover rate is the proportion of stock transactions in the market compared to the total volume of stock issued during a specific time period.
Unlike trading volume, which is challenging to directly compare across different stocks, turnover rate offers a standardized benchmark for comparing stocks with varying issuances.
This allows a direct comparison of trading activity among  different companies or markets of different sizes.
Moreover, the turnover rate provides a more precise indication of the extent of market players' activity and the liquidity of the market.
On the other hand, volume does not take into account the total number of shares issued, which makes it challenging to effectively gauge the actual liquidity of the market.
Hence, this type of specific image data format is designated as the Open-High-Low-Close Turnover-rate (OHLCT) chart.

Furthermore, the proposed approach is specifically tailored for the China A-shares stock dataset.
Within the China A-shares stock dataset, there is a notable phenomenon known as the weekend effect \citep{li2021short}.
This effect is observed in the weekly dimensionality, where the A-shares stock market tends to exhibit lower performance on Fridays and greater performance at the opening on Mondays.
This behavior might be ascribed to investors' inclination to decrease their exposure to risk prior to the weekend and thereafter re-enter the market on Mondays.
This phenomenon can result in a downward movement of stock prices on Fridays and an upward movement on Mondays.
Market-related news and events can potentially affect the weekly dimensionality.
Generally, significant corporate announcements, economic data releases, or political events take place during trading days, whereas weekends are comparatively tranquil.
Hence, in the event of noteworthy news or occurrences during the weekend, the market may see abrupt swings upon the commencement of trading on Mondays.

The weekend effect and weekly dimensionality have  significant impacts on the forecast and analysis of the A-shares stock market.
To accommodate the A-shares stock market's unique trading processes, weekends are included as time separators in the OHLCT chart.
This involves adding a column of blank pixels at the positions corresponding to the weekends.
For consistency, three black pixel columns are used as separators for rest days, aligning them with the three pixels showing the stock prices (close-open/high-low) of each day.
An advantage of this method is that it It boosts the model's capability to comprehend and internalize the recurring patterns of trade days more effectively.
As a result, this particular format for image data is designated as the Time Segmented OHLCT (TS-OHLCT) Chart.
The TS-OHLCT chart is illustrated in Figure \ref{fig1} (b).

\subsection{Proposed Methods}\label{sec3.2}

\subsubsection{Multi-scale Cascading Feature}\label{sec3.2.1}

Traditional CNN are prone to overfitting issues when employed for stock prediction \cite{jiang2023re}. The findings of their research indicate that the use of 5-day stock feature maps is more effective than the use of 20-day feature maps, which in turn is more effective than the use of 60-day feature maps. The performance gradually worsens as the number of days increases. This is due to the fact that convolutional neural networks can perform well in terms of fitting capabilities and are able to utilise any portion of local features in order to fit the labels. However, for predicting current stock price movements, the features from the most recent trading days are of particular importance. Training a CNN model with stock feature maps over longer periods can result in a locally optimal solution based on specific features. However, using too short a period of data for stock prediction can be highly risky. With a short feature time frame, it becomes difficult to understand the historical performance of the stock, making it challenging to assess its current state and trend. This, in turn, increases the risk associated with trading predictions. Therefore, we proposes a multi-scale cascade image feature approach to address the aforementioned issues.

In this feature decomposition method, a raw feature map $\mathbf{G}$ with a given time length of $n$ days can be decomposed into multiple sub-feature maps, represented by $X = \{X_1, \dots, X_C\}$, where $C$ is the number of sub-maps. Since there are 5 trading days in a week, 5 is chosen as the base for the sub-map feature window, with the calculation formula $C =\lceil \log_5(n) \rceil$. The raw feature map contains $n$ time windows, with each window corresponding to 1 day of OHLCT (Open, High, Low, Close, Turnover rate) information. Therefore, the time resolution of the raw feature map is 1 day.

Each decomposed sub-feature map $X_i$ has a time window count of 5, with a time resolution of $M_i$ days, meaning each window contains $M_i$ days of OHLCT information from the original feature map. The value of $M_i$ is determined by the following formula: For each sub-feature map $X_i$ (where \(i \in [1, C] \)), the time resolution $M_i$ is calculated as 

\begin{equation*}
    M_i = \min(5^{(i-1)}, \frac{n}{5}).
\end{equation*}

This implies that as the sub-map index $i$ increases, the number of days covered by each time window $M_i$ gradually increases, meaning the time resolution gradually decreases. In other words, sub-maps with smaller indices $X_i$ have higher time resolutions, reflecting more recent local information; whereas sub-maps with larger indices reduce the resolution by merging more days of OHLCT information, better reflecting global information. For each sub-feature map $X_i$, the feature merging for the $j$-th window (where \(j \in [1, 5]\)) can be determined using the equation below:

\begin{equation*}
\begin{aligned}
&X_{i}^{open_j} = \mathbf{G}^{Open_{(j-1) \times M_i + 1}}, \\
&X_{i}^{high_j} = \max\left(\mathbf{G}^{High_{(j-1) \times M_i + 1}}, \ldots, \mathbf{G}^{High_{j \times M_i}}\right), \\
&X_{i}^{low_j} = \min\left(\mathbf{G}^{Low_{(j-1) \times M_i + 1}}, \ldots, \mathbf{G}^{Low_{j \times M_i}}\right), \\
&X_{i}^{close_j} = \mathbf{G}^{Close_{j \times M_i}}, \\
&X_{i}^{turnover_j} = \sum_{k=(j-1) \times M_i + 1}^{j \times M_i} \mathbf{G}^{Turnover_k}.
\end{aligned}
\end{equation*}

\begin{figure}[H] 
    \centering
    \includegraphics[width=0.85\textwidth]{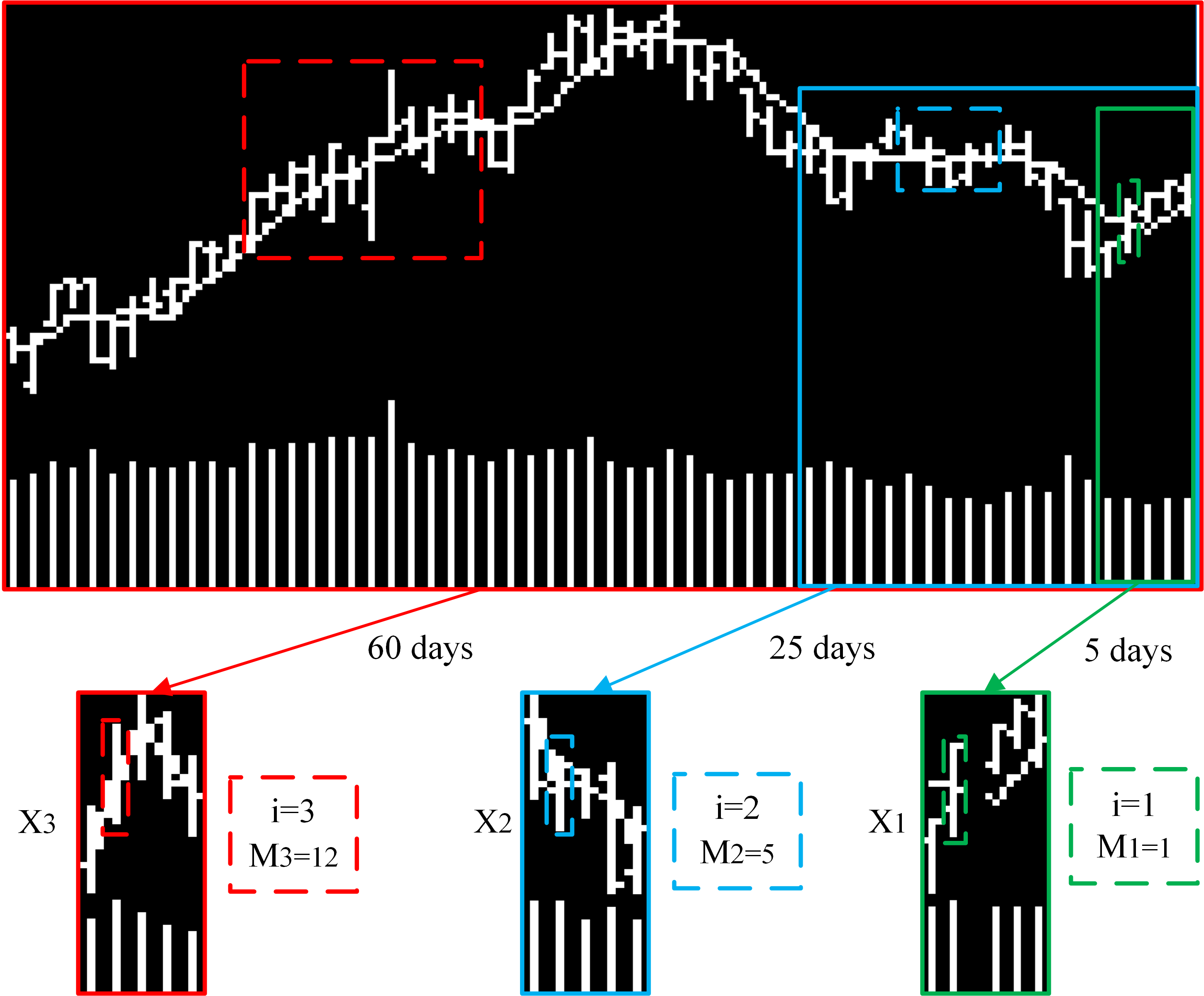}
    \caption{Illustration of the multi-scale decomposition process of a 60-day feature map. Each sub-feature map transition from high-resolution local recent data to low-resolution long-term global trends as $i$ increases.}
    \label{fig2}
\end{figure}

Figure \ref{fig2} illustrates the multi-scale decomposition process of a 60-day feature map, where the number of sub-feature maps is \( C = 3 \). The decomposed sub-feature map \( X_1 \) represents the OHLCT features of the last 5 days. This sub-feature map is treated as an independent feature map, capturing the CNN features related to recent turnover rates and price characteristics. This approach effectively constrains the convergence space of the CNN, preventing the convergence space from becoming excessively large in long-term feature maps. Simultaneously, it compresses the input of long-term features, representing them as sub-feature maps \( X_2 \) and \( X_3 \), allowing the model to learn the trend and positional information of the stock. Compared to using long-term feature maps as input, this method significantly enhances prediction accuracy while reducing network complexity.

\subsubsection{Multi-Scale Resolution (MSR) CNN}\label{sec3.2.2}

We propose a Multi-Scale Resolution (MSR) CNN to capture features from longer stock sequences. In this context, Multi-Scale Feature (MSF) modules are utilized to extract stock features from sub-feature maps at different resolution scales.
The basic building blocks of these modules include three main operations: convolution, activation, and pooling.

The convolution operation is similar to kernel smoothing, where it scans both horizontally and vertically across the image to extract features from each element in the image matrix, and thus generating contextual features.
The output of the convolution filters is then processed through an activation function, specifically using the Leaky ReLU activation function, which introduces non-linearity.
The final operation in the network module is max pooling, which performs another round of scanning over the input matrix and returns the maximum value from adjacent regions in the image.
This process effectively reduces the dimensionality of the data and minimizes noise, thereby generating a high-dimensional feature set.

\begin{figure}[H] 
    \centering
    \includegraphics[width=0.85\textwidth]{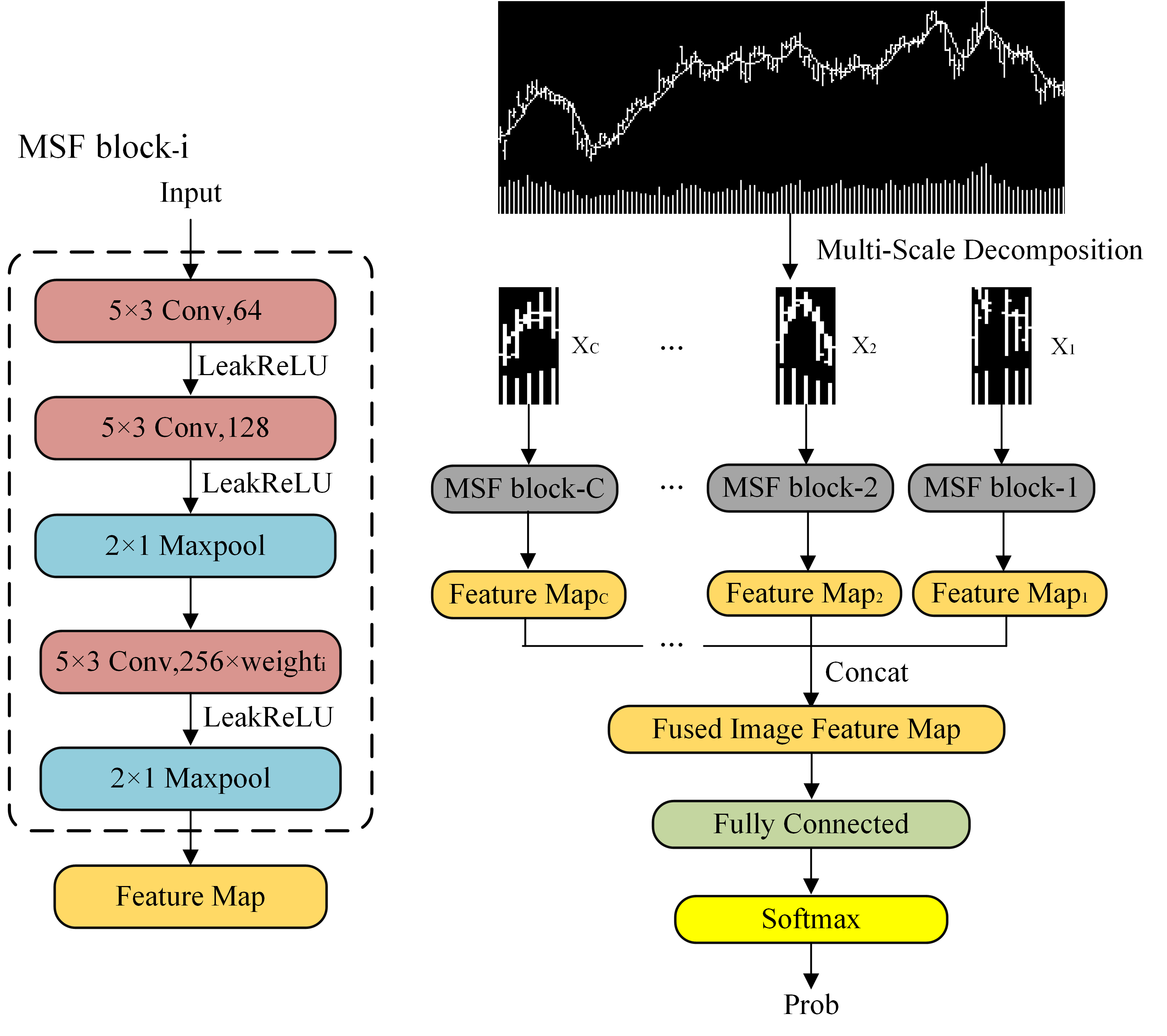}
    \caption{Our proposed network architecture of MSR-CNN.}
    \label{fig3}
\end{figure}

After these operations, features from different scales are merged, and a fully connected layer is added.
This fully connected layer is activated using the Softmax function to produce the prediction results.
The primary goal of the proposed MSR-CNN is to perform binary classification, specifically to predict whether the value of a given stock will rise (labeled as 1) or fall (labeled as 0) over a specified period.
As such, the CNN's prediction can be interpreted as an estimate of the probability of positive returns.
The network architecture is shown in Figure \ref{fig3}.

As shown in Figure \ref{fig3}, we first decompose the stock feature map with a time length of \(n\) into \(C\) sub-feature maps at different resolution scales.
Each sub-feature map has a different feature weight \( \text{weight}_i \) (where \(i \in [1, C]\)). As \(i\) increases, \(X_i\) contains stock features from further back in time, and so the designed feature weights gradually decrease.
This approach allows the model to focus more on recent local features.
The formula for the feature weights is as follows:
\begin{equation*}
\text{weight}_i = 
\begin{cases} 
0.5, & \text{if } i = 1, \\
\frac{0.5}{2^{i-2}}, & \text{if } 2 \leq i \leq C-1, \\
\frac{0.5}{2^{C-2}}, & \text{if } i = C.
\end{cases}
\end{equation*}
Taking \(C=3\) as an example, our feature weighted vector's \( \text{weight} \) is $([0.5, 0.25$, \\ $ 0.25])$, which indicates that the sub-feature map \(X_1\) representing the most recent 5 days has a weight proportion of 50\% (i.e., the MSF block-1 will extract a 128-dimensional feature map).
The sub-feature map \(X_2\), which compresses the recent 25 days, has a feature weight of 25\%, and the global feature map \(X_3\) also has a feature weight of 25\%.

Finally, we integrate the multi-scale features extracted by each MSF block \(i\) to capture stock trend information across different time scales.
The integrated features are processed through a fully connected layer and the probabilities of stock price movements are obtained using the Softmax function.
This approach comprehensively leverages multi-scale features and addresses the issue of local overfitting that CNNs encounter as stock feature maps grow \citep{jiang2023re}.
Our experimental results demonstrate that the prediction accuracy of the multi-scale resolution network is significantly higher than that of the single-scale network.

\subsubsection{Sequence Multi-Scale Fusion Regression (SMSFR) CNN}\label{sec3.2.3}

Although MSR-CNN achieves high classification accuracy, in the context of the stock market, our primary focus is often on stock returns.
Specifically, we are interested in predicting the magnitude of the future price changes of a particular stock, rather than just predicting whether it will rise or fall.
However, traditional CNN models based on pixel-level data are not suitable for directly predicting percentage changes in stock prices.
Applying MSR-CNN directly to the regression tasks is insufficient to model the complexities of the stock market.

Although using MSR-CNN to predict percentage changes in the field of stock trading is challenging, regression labels offer several advantages over classification labels.
Regression labels provide a comprehensive perspective on stock behavior by introducing continuous data features such as price and trading volume, contributing to more accurate predictions of dynamic stock market.
They generate continuous numerical outputs, capture the inherent volatility of the stock market and support  advanced quantitative risk assessments, thereby facilitating more effective portfolio management.

To fully leverage the potential of sequence CNN in predicting stock market movements, we further propose the Sequence Multi-Scale Fusion Regression (SMSFR) CNN, which combines both classification and regression labels for stock prediction.
Without altering the original MSR-CNN architecture, we treat the sequence dimension of stocks as an independent path, extracting sequence features through convolution, then fusing them with image features, and predicting stock price movements through fully connected layers.
The time series features consist of 12 features over 30 trading days.
These features include six original data features ($close$, $open$, $high$, $low$, $ma5$, $turnover rate$), two time features ($month$ and $week$), and four extended price features ( $close_{ratio}$, $open_{ratio}$, $high_{ratio}$, $low_{ratio}$).
Moving Average 5 ($ma5$) is a technical analysis indicator that represents the average of the last 5 closing prices of an asset, helping to smooth out price data and identify trends.
The extended features are calculated as follows:
\begin{equation*}
\footnotesize
\text{ratio}_{t} = 
\begin{pmatrix}
\text{close}_{t} \\
\text{open}_{t} \\
\text{high}_{t} \\
\text{low}_{t}
\end{pmatrix}
-
\begin{pmatrix}
\text{close}_{t-1} \\
\text{close}_{t-1} \\
\text{close}_{t-1} \\
\text{close}_{t-1}
\end{pmatrix}
\Bigg/ 
\begin{pmatrix}
\text{close}_{t-1} \\
\text{close}_{t-1} \\
\text{close}_{t-1} \\
\text{close}_{t-1}
\end{pmatrix}
= 
\begin{pmatrix}
\frac{\text{close}_{t} - \text{close}_{t-1}}{\text{close}_{t-1}} \\
\frac{\text{open}_{t} - \text{close}_{t-1}}{\text{close}_{t-1}} \\
\frac{\text{high}_{t} - \text{close}_{t-1}}{\text{close}_{t-1}} \\
\frac{\text{low}_{t} - \text{close}_{t-1}}{\text{close}_{t-1}}
\end{pmatrix}
=
\begin{pmatrix}
\text{close}_{\text{ratio}} \\
\text{open}_{\text{ratio}} \\
\text{high}_{\text{ratio}} \\
\text{low}_{\text{ratio}}
\end{pmatrix}.
\end{equation*}

To eliminate the dimensional differences between various features, we normalize each segment of the input time series data \(X \in \mathbb{R}^{30 \times 12}\), where each row represents the feature vector for a single day.
Since the four extended price features already include normalization in their calculations, we only normalize the remaining six original features and the two time features to maintain consistency in the relative scale of the data.
The normalization process for the extended features is as follows:
\begin{equation*}
{X_{i}^{j}}' = \frac{X_{i}^{j}}{X_{30}^{\text{close}}}, \quad \forall i \in \{1, 2, \dots, 30\}, j \in \{\text{close}, \text{open}, \text{high}, \text{low}, \text{ma5} \},
\end{equation*}
where \(X_{30}^{close}\) represents the $close$ value of the last day in the input sequence.
Additionally, the time dimensions $month$ and $week$ are normalized to ensure consistent scaling.
The normalization process for these dimensions is as follows:

\[
\text{month}_{\text{normalized}} = \frac{\text{month}}{12}, \quad \text{week}_{\text{normalized}} = \frac{\text{week}}{5}.
\]

\begin{figure}[ht]
    \centering
    \includegraphics[width=\textwidth]{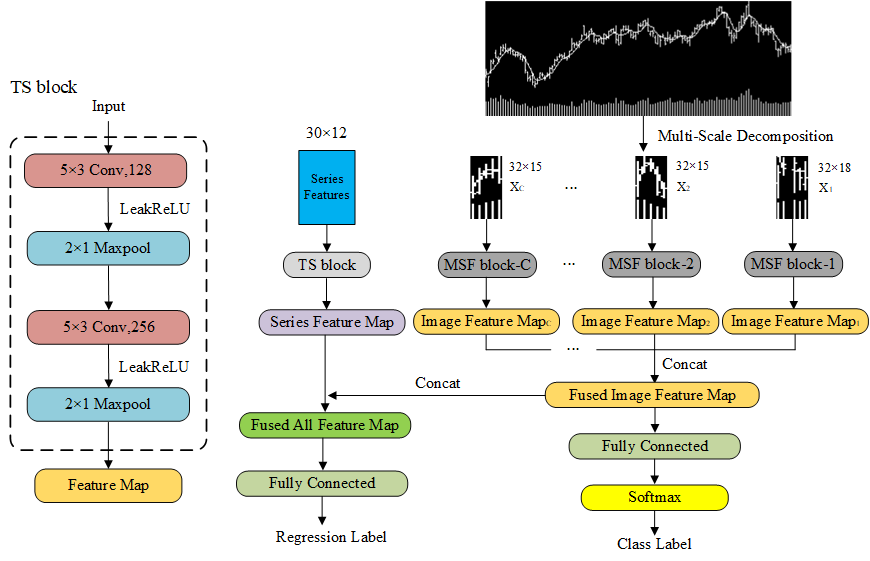}
    \caption{Our proposed network architecture of SMSFR-CNN.}
    \label{fig4}
\end{figure}

By normalizing the $month$ by 12 and the $week$ by 5, we convert these features into a relative scale, where the range of values between 0 and 1.
This approach ensures that the time-related features are consistent with other normalized features in the dataset.
The SMSFR-CNN network architecture is illustrated in Figure \ref{fig4}.

The TS block (Timeseries block) is a module designed for extracting features from time-series data. It consists of a 5×3 convolutional layer with 128 channels, followed by a Leaky ReLU activation function to capture temporal features. After that, a 2×1 max-pooling layer is applied for downsampling. A second 5×3 convolutional layer with 256 channels is used to extract more complex features, again followed by a Leaky ReLU activation. Finally, another 2×1 max-pooling layer is used for further downsampling, and the output is a feature map representing the extracted time-series features.

Another advantage of using sequence feature embedding in the model is that during the training process, when updating the parameters, the Mean Squared Error (MSE) of the regression labels is propagated backward to update the parameters of the MSF block in the image part.
We believe that this effectively adds more stock return information to the image CNN modules while constraining the convergence space and direction of image features.
As a result, the training process becomes more stable, further enhancing the model excess annualized return.

\section{Experiments}\label{sec4}

\subsection{China A-shares Stock Data}\label{sec4dataset}

The Science and Technology Innovation Board (STAR Market) and Beijing Stock Exchange stocks are characterized by small market capitalization, having been listed after July 2019, and are subject to various trading restrictions.
In our experiments, the dataset encompasses all stocks in the China A-share stock market, excluding those from the STAR Market (codes starting with 688*) and the Beijing Stock Exchange (codes starting with 8*), totaling 4,454 stocks.

\begin{table}[h]
    \centering
    \caption{Dataset composition and split details for the China A-share stock market in our experiments.}\label{tab.dataset}
    \begin{tabularx}{\linewidth}{lllc}
        \hline
        \hline
        \textbf{Dataset} & \textbf{From} & \textbf{To} & \textbf{Counts} \\
        \hline
        \hline
        Training Set & January 1, 2012 & December 31, 2020 & 5,958,162 \\
        Validation Set & January 1, 2021 & December 31, 2021 & 754,893 \\
        Test Set & January 1, 2022 & June 30, 2024 & 1,907,335 \\
        \hline
        Total & January 1, 2012 & June 30, 2024 & 8,620,390 \\
        \hline
        \hline
    \end{tabularx}
\end{table}

Daily trading data for these 4,454 stocks were collected from January 1, 2012, to June 30, 2024, using the Tonghuashun API, a prominent stock data API platform in China.
Each record contains several features of the respective stock, including numerical data such as opening price, closing price, highest price, lowest price, trading volume, and turnover rate, as well as two binary features indicating whether the stock price increased or decreased (represented by 1 or 0, respectively).
A notable feature of the Chinese A-share market is the price limit mechanism, which prohibits trading when a stock's price reaches its upper limit on the preceding trading day.
Consequently, in our data preprocessing phase, we exclude the samples where the final candlestick within the specified window exhibited an upper limit price movement.
The dataset is divided into three chronological subsets: a training set encompassing 5,958,162 observations from January 1, 2012, to December 31, 2020; a validation set comprising 754,893 observations from January 1, 2021, to December 31, 2021; and a test set containing 1,907,335 observations from January 1, 2022, to June 30, 2024.
Table \ref{tab.dataset} presents the split details of our dataset.

\subsection{Classification Experiments}\label{sec4.2}
\subsubsection{Experimental Setup}\label{sec4.1}

We first evaluate the accuracy of different methods in predicting stock price movements.
This task can be viewed as a binary classification problem, where the classes are typically divided into positive and negative categories with the corresponding samples referred to as positive and negative examples.
When a stock is expected to rise in the future, the label is defined as $y=1$. 
Otherwise, if the stock is expected to fall, the label is defined as $y=0$.
During the training process for classification problems, we use the cross-entropy loss function to minimize the standard objective function.
The cross-entropy loss function is defined as follows:

\begin{equation*} 
\label{eq5} 
L(y,\hat{y}) = -y\log(\hat{y}) - (1-y)\log(1-\hat{y}),
\end{equation*}
where $\hat{y}$ represents the output from the softmax function in the final layer of the model.
We select  the Positive Predictive Value (PPV) and Negative Predictive Value (NPV) as the metrics for evaluation, which are defined as follows:

\begin{equation*} 
\label{eq7} 
PPV = \frac{TP}{TP + FP}, \quad NPV = \frac{TN}{TN + FN}, 
\end{equation*}
where $TP$ and $FP$ denote the true positives and false positives, respectively.
$TN$ and $FN$ denote the true negatives and false negatives, respectively.

To eliminate experimental bias caused by random seeds and stochastic gradient descent, we conduct  10 repeated training processes for each subsequent method.
In each training process, we apply an early stopping method after the 5th epoch and selected the model with the best performance on the validation set.
Finally, we take the average of the results from the 10 testing to reflect the overall performance of the method.

Additionally, to prevent early overfitting during the training process, we employ the regularization procedure mentioned in \citep{gu2020empirical}.
In each layer, we apply the Xavier weight initializer \citep{glorot2010understanding} to ensure that the variance of the initial weights is comparable to the variance of the labels, thereby promoting faster convergence of the model.
The optimization of the loss function was carried out using stochastic gradient descent and the Adam algorithm \citep{kingma2015adam}, with an initial learning rate set to $3\times 10^{-5}$ and a batch size of 256.
All experiments were conducted on 8 Tesla V100 GPUs with 32GB memory each (a total of 256GB) and 80-core CPUs.

\subsubsection{Comparison Between Time Series and Image Models on Original Stock Data}
We select  the original stock features with duration  of 5 days, 20 days, and 60 days to predict the probability of stock price movements over the next 5 days.
The input features include basic price and volume indicators such as open, high, low, close, $ma5$, and turnover rate.
Additionally, we substitute  the trading volume with the turnover rate and introduced time delimiters to verify whether these factors contribute to improved stock prediction.

In this experiment, we choose Timemixer \citep{wangtimemixer}, Timesnet \citep{wu2022timesnet}, Dlinear \citep{zeng2023transformers}, and PatchTST \citep{nie2022time} as the baseline models for time series analysis.
Additionally, we select the OHLCV+CNN model \citep{jiang2023re} as the baseline model for image-based analysis.
This experimental design aims to validate whether the time series models can capture more information as the input time length increases and to assess whether the image-based models are prone to overfitting.
The results of the experiment are displayed in Table \ref{tab.baseline}.

\begin{table}[H]
\caption{Performance comparison of time series baselines, the proposed OHLCT data format, and the TS-CNN model for stock price prediction with varying input lengths (5, 20, and 60 days).} \label{tab.baseline}
\centering
\scriptsize
\begin{tabular}{lccccc}
\hline
\hline
&\textbf{Method}& \textbf{Input Lengths} & \textbf{Predict Lengths} & \textbf{PPV}(\%)$\uparrow$& \textbf{NPV}(\%)$\uparrow$ \\
\hline
\hline
\multirow{12}{*}{\textbf{Series}}&Timemixer \citep{wangtimemixer} & 5 days & 5 days & $52.36$ & $54.53$ \\
&Timemixer  & 20 days & 5 days & $54.26$ & $57.34$ \\
&Timemixer  & 60 days & 5 days & \textbf{56.58} & \textbf{59.15} \\
\cline{2-6}
&TimesNet \citep{wu2022timesnet} & 5 days & 5 days & $52.09$ & $54.39$ \\
&TimesNet & 20 days & 5 days & $54.00$ & $57.10$ \\
&TimesNet  & 60 days & 5 days & $55.94$ & $58.87$ \\
\cline{2-6}
&DLinear \citep{zeng2023transformers} & 5 days & 5 days & $52.58$ & $54.26$ \\
&DLinear & 20 days & 5 days & $53.68$ & $56.72$ \\
&DLinear  & 60 days & 5 days & $55.91$ & $58.96$ \\
\cline{2-6}
&PatchTST \citep{nie2022time} & 5 days & 5 days & $52.16$ & $54.82$ \\
&PatchTST& 20 days & 5 days & $53.91$ & $56.82$ \\
&PatchTST& 60 days & 5 days & $56.31$ & $58.76$ \\
\hline
\hline
\multirow{9}{*}{\textbf{Image}}&OHLCV+CNN \citep{jiang2023re} & 5 days & 5 days & \textbf{55.32} & \textbf{58.12} \\
&OHLCV+CNN  & 20 days & 5 days & $54.88$ & $57.76$ \\
&OHLCV+CNN  & 60 days & 5 days & $54.64$ & $57.73$ \\
\cline{2-6}
&OHLCT+CNN  & 5 days & 5 days & \textbf{55.47} & \textbf{58.13} \\
&OHLCT+CNN  & 20 days & 5 days & $55.01$ & $57.91$ \\
&OHLCT+CNN  & 60 days & 5 days & $54.68$ & $57.82$ \\
\cline{2-6}
&TS-CNN(ours)  & 5 days& 5 days& \textbf{56.08} & \textbf{58.54} \\
&TS-CNN(ours)  & 20 days& 5 days & $55.89$ & $58.12$ \\
&TS-CNN(ours)  & 60 days& 5 days & $55.34$ & $57.87$ \\
\hline
\hline
\end{tabular}
\end{table} 

From the results shown in Table \ref{tab.baseline}, it can be analyzed that as the input duration for stocks increases, the time series-based models are able to learn more historical information about the stocks, thereby improving their ability to predict the trend of individual stocks over the next 5 days.
The experimental results indicate that the time series-based models will not overfit and achieve better performance than CNN-based methods with the input length increases.
In contrast, the CNN model based on image features tends to overfit to local features, leading to prediction performance in the order of 5 days $>$ 20 days $>$ 60 days.
Additionally, according to all experiments, it is obviously that replacing the volume with the turnover ratio in the predictions yields slightly better results than the original OHLCV+CNN.
The CNN models incorporating time-split features show a noticeable improvement in accuracy, but they still exhibit overfitting to image features.

\subsubsection{Multi-Scale Feature Decomposition on Stock Data}

We further compare a cascaded CNN architecture based on multi-scale decomposition, decomposing the image features of 20 days and 60 days, and inputting the sub-feature maps into the network according to their weights.
The input of the first method, MSR, consists purely of image features, with experimental hyperparameters and loss functions consistent with those of the Original Stock Data Experiment. The second method, SMSFR, includes both multi-scale image features and sequence features in its input.
The prediction labels of the sequence part are also treated as a regression problem.
The rise and fall amplitude of the stock 5 days later (up or down) is used as the label $y$, consistent with the time period of the classification labels. The training loss function for regression is defined as the mean squared error (MSE) loss, which is defined as follows:

\begin{equation*} \label{eq6}
L(y,\hat{y})=\frac{1}{n} \sum_{i=1}^{n} (y_i-\hat{y_i})^2,
\end{equation*}
where $\hat{y}$ is the output of the final layer of the model.
By minimizing the MSE loss function, we aim to make the predicted values as close as possible to the actual stock price changes. Finally, we still use the 10-experiment scheme from the Original Stock Data Experiment, averaging the test set results of the best-performing model on the validation set across 10 repeated experiments. (Here, we introduce regression labels merely to enable the image features to learn more information about stock price changes and to constrain the solution space through backpropagation, thus still using classification labels as the criterion for method evaluation.)

\begin{table}[H]
\caption{Performance comparison of time series baselines, our proposed MSR, and SNSFR models for stock price prediction. Only 20 and 60 days are considered in this experiment.}
\label{tab:smsfr}
\centering
\footnotesize
\begin{tabular}{ccccc}
\hline
\hline
\textbf{Method}& \textbf{Input Lengths} & \textbf{Predict Lengths} & \textbf{PPV}(\%)$\uparrow $& \textbf{NPV}(\%)$\uparrow $\\
\hline
\hline
Timemixer \citep{wangtimemixer} & 20 days & 5 days & $54.26$ & $57.34$ \\
TimesNet \citep{wu2022timesnet} & 20 days & 5 days & $54.00$ & $57.10$ \\
DLinear \citep{zeng2023transformers} & 20 days & 5 days & $53.68$ & $56.72$ \\
PatchTST \citep{nie2022time} & 20 days & 5 days & $53.91$ & $56.82$ \\
OHLCV+CNN\citep{jiang2023re} & 20 days& 5 days & $54.88$ & $57.76$\\
MSR(ours)  & 20 days& 5 days & $58.68$ & $60.33$\\
\textbf{SMSFR(ours)}  &  20 days& 5 days & \textbf{60.41} & \textbf{62.53} \\
\hline
Timemixer& 60 days & 5 days & $56.58$ & $59.15$ \\
TimesNet & 60 days & 5 days & $55.94$ & $58.87$ \\
DLinear  & 60 days & 5 days & $55.91$ & $58.96$ \\
PatchTST & 60 days & 5 days & $56.31$ & $58.76$ \\
OHLCV+CNN\citep{jiang2023re} & 60 days& 5 days & $54.64$ & $57.73$\\
MSR(ours)  & 60 days& 5 days & $59.53$ & $61.25$ \\
\textbf{SMSFR(ours)}  & 60 days& 5 days & \textbf{61.15} & \textbf{63.37} \\
\hline
\hline
\end{tabular}
\end{table} 

Since the 5-days stock feature is the smallest unit of the sub-feature map, TS-CNN 5-days can be considered equivalent to MSR 5-days.
Hence, 5-days stock feature is not considered in all proposed decomposed-based methods.
Experimental results indicate that the accuracy of both multi-scale feature decomposition methods increases as the input duration of the stock data is extended. This suggests that decomposition-based methods are better at capturing both the short-term local features and long-term trend features of stocks, thereby addressing the issue of local overfitting in the original image features.

Meanwhile, the Sequence Multi-Scale Fusion Regression (SMSFR) CNN achieved the best results when the input features spanned 60 days, with an accuracy of 61.15\% for predicting stock price increases and 63.37\% for predicting decreases after 5 days. This indicates that integrating sequential features into image features, along with incorporating stock price fluctuation labels, allows the model to gain a deeper understanding of the underlying dynamics of stocks. Since the fluctuation magnitude reflects the strength and volatility characteristics of individual stocks, which cannot be captured by simple classification labels, the model is better equipped to grasp the principles underlying stock price movements, thereby enhancing prediction accuracy.

\subsection{Profitability Backtest Experiments}

\subsubsection{Experimental Setup}
To better evaluate the performance of the models in the A-share market, we conducte  a simulated backtest for a two-and-a-half-year period from January 1, 2022, to June 30, 2024, using all the methods described above.
For each method, we select the model that performed best on the validation set across 10 repeated experiments.
To simulate real trading conditions, we set the maximum open positions for each model to 5.
When the model predicts a stock's probability of rising to be greater than 80\%, we buy  the stocks in descending order of probability and sell them after 5 trading days.
To account for the slippage effects in actual trading signals, we calculate a transaction cost of 0.3\%.
The final result for each method is the average of the simulation results of the 10 corresponding models.
We use $PF$ to represent total profit and max drawdown (MDD).
The results of the experiment are displayed in Table \ref{tab:tab333}.

\begin{table}[H]
\caption{Profit and max drawdown comparison of time series baselines, reproduced OHLCV-based CNN, our propose MSR, and SNSFR models for stock price prediction with different input lengths.}\label{tab:tab333}
\centering
\scriptsize
\begin{tabular}{ccccc}
\hline
\hline
\textbf{Method} &  \textbf{Input Lengths} & \textbf{Predict Lengths} & \textbf{PF(Ave)}(\%)$\uparrow$& \textbf{MDD(Ave)}(\%)$\downarrow$ \\
\hline
\hline
Timemixer  & 5 days & 5 days & $13.52$ & $35.78$ \\
TimesNet   & 5 days & 5 days & $4.27$ & $32.67$ \\
DLinear  & 5 days & 5 days & $9.97$ & $35.04$ \\
PatchTST & 5 days & 5 days & $5.36$ & $39.07$ \\
\hline
Timemixer & 20 days & 5 days & $25.47$ & $28.99$ \\
TimesNet  & 20 days & 5 days & $21.17$ & $31.66$ \\
DLinear  & 20 days & 5 days & $19.98$ & $31.92$ \\
PatchTST & 20 days & 5 days & $19.62$ & $29.74$ \\
\hline
Timemixer  & 60 days & 5 days & $32.17$ & $32.62$ \\
TimesNet   & 60 days & 5 days & $30.51$ & $31.86$ \\
DLinear  & 60 days & 5 days & $26.56$ & $30.11$ \\
PatchTST & 60 days & 5 days & $29.56$ & $32.51$ \\
\hline
\hline
OHLCV+CNN & 5 days & 5 days & $23.15$ & $29.72$ \\
OHLCV+CNN & 20 days& 5 days & $22.53$ & $29.42$ \\
OHLCV+CNN & 60 days& 5 days & $18.96$ & $30.46$ \\
\hline
MSR(ours)  & 20 days& 5 days & $83.04$ & $31.46$ \\
MSR(ours) & 60 days& 5 days & $108.48$ & $32.77$ \\
\hline
\textbf{SMSFR(ours)}  & 20 days& 5 days & \textbf{144.49} & \textbf{28.35} \\
\textbf{SMSFR(ours)}  & 60 days& 5 days & \textbf{165.09} & \textbf{27.92} \\
\hline
\hline

\end{tabular}
\end{table}
The experimental results demonstrate that our proposed MSR and SMSFR methods significantly improve the model's returns. The SMSFR-based method introduces regression labels, which limit the model's convergence space and achieve the smallest maximum drawdown and the best returns across different time feature inputs. The best result is achieved by the 60-day SMSFR, which yielded a return of 165.09\% and a maximum drawdown of 27.92\% during the period from January 1, 2022, to June 30, 2024, significantly outperforming other methods.

\begin{table}[hp]
\caption{Different indexes between January 1, 2012 and June 30, 2024 for IDC and IMD.}\label{tab:tab444}
\centering
\small
\begin{tabular}{ccccc}
\hline
\hline
\textbf{Index} &  \textbf{From} & \textbf{To} & \textbf{IDC}(\%)& \textbf{IMD}(\%) \\
\hline
\hline

\textbf{SSE Index}  &January 1, 2012 & June 30, 2024 & -18.31 & 25.61 \\
\textbf{CSI 300 Index}  & January 1, 2012 & June 30, 2024 & -29.61 &35.35 \\
\textbf{ChiNext Index}  & January 1, 2012 & June 30, 2024 & -48.21 & 52.30 \\
\hline
\hline
\end{tabular}
\end{table}

In Table \ref{tab:tab444}, to more intuitively compare the backtesting performance of each method, we select the two most representative indices of the A-share market: the Shanghai Stock Exchange Composite Index (SSE Index) and the China Securities Index 300 (CSI 300 Index), as well as the ChiNext Composite Index (ChiNext Index), which best reflects the real market environment.

The SSE Index is a comprehensive index that reflects the overall performance of companies listed on the Shanghai Stock Exchange, covering all stocks listed on the Shanghai Stock Exchange. It represents the overall trend of large-cap and small-cap stocks in the A-share market.

The CSI 300 Index is compiled by the China Securities Index Company and consists of the 300 largest and most liquid stocks from the Shanghai and Shenzhen stock exchanges. This index aims to reflect the overall performance of blue-chip stocks in the Chinese A-share market.

\begin{figure*}[htp]
	\centering
	\includegraphics[scale=0.5]{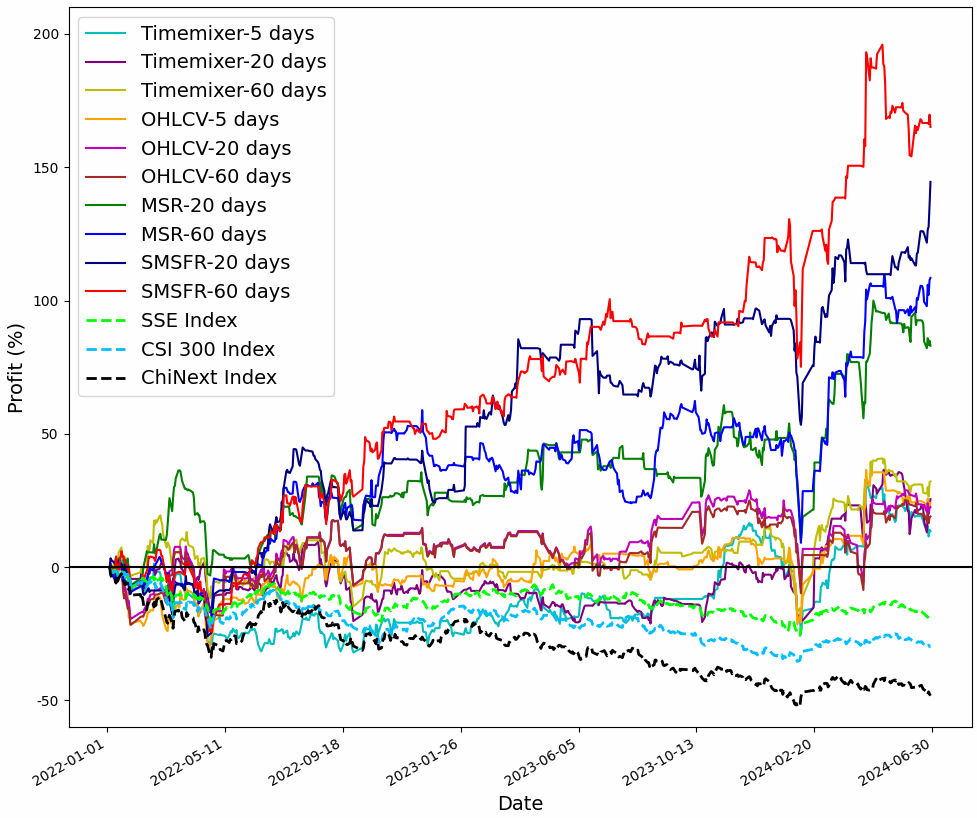}
	\caption{Profit comparison among time series baselines, reproduced OHLCV-based CNN, our proposed MSR and SNSFR models, and three market indexes on the test set.}
	\label{fig.profit}
\end{figure*}

The ChiNext Index is a comprehensive index of the ChiNext market of the Shenzhen Stock Exchange, covering all stocks listed on the ChiNext market. This index mainly reflects the overall performance of emerging industries and innovative enterprises in China, with high growth and volatility, and is therefore considered a key indicator for measuring the performance of the ChiNext market in China.

By comparing the performance of our methods with these benchmark indices, we can more comprehensively evaluate their backtest effectiveness under different market conditions. We use Index Change (IDC) and Index Max Drawdown (IMD) to assess the performance of the indices.

From the Figure \ref{fig.profit}, it can be observed that various strategies significantly outperform the benchmark indices, with different strategies showing varying strengths and weaknesses across different time windows.
Experimental results show that as the input duration of the stock data increases, the returns of multi-scale feature decomposition methods significantly exceed those of other methods.
This suggests that multi-scale image decomposition-based methods perform better in capturing both short-term local features and long-term trend features of stocks. Moreover, the results of the SMSFR method, which incorporates time-series features, outperform those of the purely image-based MSR method.
In particular, when the input features span 60 days, the SMSFR method achieved the optimal results, with an absolute return of 165.09\% during the period from January 2022 to June 30, 2024, far surpassing the performance of the three major A-share market indices.

\section{Conclusions}\label{sec5}
In this study, we apply temporal separators to image features, successfully addressing the issue of traditional stock feature maps failing to capture holiday information in the A-share market. Additionally, we propose a novel multi-scale image decomposition method aimed at solving the problems of overfitting in image features for stock price prediction and the difficulty of capturing the complexity of temporal dynamics in financial data.

The experimental data indicates our proposed Multi-Scale Resolution CNN (MSR) effectively prevents the model from falling into local overfitting when handling long time-series features by utilizing the multi-scale features of stock image representations. By decomposing long-sequence stock features into sub-feature maps at different time scales, our model can simultaneously capture short-term fluctuations and long-term trends in stock prices, thereby significantly improving prediction accuracy.

A key contribution of this study is the integration of regression labels into the convolutional neural network framework, further proposing the Sequence Multi-Scale Fusion Regression CNN (SMSFR). This approach can  learns and predicts the magnitude of stock price changes better, which is particularly important in financial markets where the magnitude of price fluctuations is often more critical than the direction. By combining image feature extraction with sequential data, our model can comprehensively understand the complex relationships between different time periods and trading patterns, achieving optimal prediction results.

However, this study also has some limitations.
Firstly, the stock market is an extremely complex system, and many traditional methods are prone to failure in practical applications, especially when market dynamics change dramatically, making predictions more challenging.
Currently, most deep learning methods are based on time series data, while the development and application of image features are relatively rare.
Although we have proposed a solution for image features in this study, its long-term profitability and robustness have not been fully validated.
As more image-related research progresses, the practical application effects of image feature methods and the sustainability of their profitability in the stock market still require further in-depth investigation.

Despite the encouraging results, there are several avenues for future work. First, this study did not evaluate the regression labels in the SMSFR. The role of regression labels in this work is to introduce additional information about stock price fluctuations to help the image feature component converge, enabling the model to better understand stock volatility and underlying trends. In future work, we will use regression labels as an important factor in evaluating stock prediction performance. Additionally, exploring other deep learning architectures, such as attention mechanisms or graph neural networks, may further improve the capture of complex dependencies in stock data.

The SMSFR-CNN model has made significant progress in stock price prediction by effectively integrating multi-scale temporal features and sequential data into a unified framework. This approach enhances the interpretability of predictions and also provides more accurate and reliable forecasts, which are of great value to investors and financial analysts. Future research will focus on evaluating the model's prediction of stock price fluctuations, expanding the model architecture, and exploring its application in other financial markets.

\section*{Acknowledgements}
This work is supported by the 
the grants from the National Natural Science Foundation of China (12371023,12271338), and the Natural Sciences and Engineering Research Council of Canada (NSERC) (RGPIN 2020-06746), the joint research and Development fund of Wuyi University, Hong Kong and Macao (2019WGALH20). Experimental Teaching Demonstration Center for Intelligent Financial Talents, Beijing Institute of Technology, Zhuhai (2023006ZLGC), and the Research Base for Intelligent Financial Governance and National Economic Security, Beijing Institute of Technology, Zhuhai (2023WZJD009).

\bibliographystyle{elsarticle-num} 
\bibliography{bibliography}
\end{document}